\documentclass{article}

\usepackage{arxiv}

\usepackage[utf8]{inputenc} 
\usepackage[T1]{fontenc}    
\usepackage{hyperref}       
\usepackage{url}            
\usepackage{booktabs}       
\usepackage{amsfonts}       
\usepackage{nicefrac}       
\usepackage{microtype}      
\usepackage{lipsum}
\usepackage{amsmath}
\usepackage{amsbsy}
\usepackage{bm}

\usepackage[dvipsnames]{xcolor}

\newcommand{\vect}[1]{\bm{#1}}
\newcommand{\mat}[1]{\vect{\mathrm{#1}}}
\newcommand{\norm}[1]{\left\lVert#1\right\rVert}

\newcommand{\warmedUpThreshold}{L}

\newcommand{\cZero}{\vect{c}_0}
\newcommand{\hZero}{\vect{h}_0}

\newcommand{\utSeq}{\{u_t\}}

\newcommand{\xdata}{\{\vect{x}_i\}}
\newcommand{\embedding}{\vect{\phi}}

\newcommand{\embeddingMatrix}{\mat{\Phi}}
\newcommand{\manifold}{\Phi}
\newcommand{\manifoldch}{\Psi}
\newcommand{\secondKernel}{C}

\newcommand{\Fproj}{\tilde{F}}
\newcommand{\new}{\mathrm{new}}
\newcommand{\phinew}{\vect{\phi}_{\new}}

\newcommand{\cell}{\mathbf{c}}
\newcommand{\hidden}{\mathbf{h}}
\newcommand{\predictorNN}{g}

\newcommand{\sect}{Sec.}
\newcommand{\fig}{Fig.}

\usepackage{graphicx}

\usepackage{tikz}
\usetikzlibrary{positioning, fit}

\title{Initializing LSTM internal states \\
via manifold learning}

\author{
  Felix P. Kemeth\\
  Department of Chemical and Biomolecular Engineering\\
  Whiting School of Engineering, Johns Hopkins University\\
  3400 North Charles Street, Baltimore, MD 21218, USA\\
  \texttt{fkemeth1@jhu.edu} \\
  \AND
  Tom Bertalan\\
  Department of Chemical and Biomolecular Engineering\\
  Whiting School of Engineering, Johns Hopkins University\\
  3400 North Charles Street, Baltimore, MD 21218, USA\\
  \AND
  Nikolaos Evangelou\\
  Department of Chemical and Biomolecular Engineering\\
  Whiting School of Engineering, Johns Hopkins University\\
  3400 North Charles Street, Baltimore, MD 21218, USA\\
  \AND
  Tianqi Cui\\
  Department of Chemical and Biomolecular Engineering\\
  Whiting School of Engineering, Johns Hopkins University\\
  3400 North Charles Street, Baltimore, MD 21218, USA\\
  \AND
  Saurabh Malani\\
  The Department of Chemical and Biological Engineering \\
  Princeton University\\
  Princeton, NJ 08544, USA\\
  \AND
  Ioannis G. Kevrekidis\\
  Department of Chemical and Biomolecular Engineering\\
  Whiting School of Engineering, Johns Hopkins University\\
  3400 North Charles Street, Baltimore, MD 21218, USA\\
  \texttt{yannisk@jhu.edu}
}

\begin{document}
\maketitle

\begin{abstract}
  We present an approach, based on learning an intrinsic data manifold, for the initialization of the internal state values
  of LSTM recurrent neural networks, ensuring consistency with the 
  initial observed input data.
  Exploiting the generalized synchronization concept, we argue that the converged, ``mature"  internal states constitute a function on this learned manifold.
  The dimension of this manifold then dictates the length of observed input time series data required for consistent initialization.
  We illustrate our approach through a partially observed chemical model system, where initializing the internal LSTM states in this fashion yields visibly improved performance.
  Finally, we show that learning this data manifold enables the transformation of partially observed dynamics into fully observed ones, facilitating alternative identification paths for nonlinear dynamical systems.
\end{abstract}

\section{Introduction}
\label{sec:introduction}

Sequence modeling has become one of the fundamental tasks
in many applied and scientific disciplines.
Examples include audio and speech,
as well as engineering, robotics and finance, to name a few.
The challenges in these fields typically involve tasks such as prediction,
control or natural language processing (NLP).
This need has established a long-standing effort to find ways to model
the dynamics of such temporal data. A particular class of functions demonstrated to excel in this task are recurrent neural networks (RNNs).

One type of such RNNs are sliding-window approaches (e.g. ~\cite{osti_5470451, HUDSON19902075, gers01_applying_lstm_time_series_pred}.)
Examples range from time-delay neural networks~\cite{haffner92_multi_step_time_delay}
and neural networks templated on explicit integrators~\cite{rico-martinez92_discr_vs},
to nonlinear autoregressive models with exogenous inputs (NARX)~\cite{lin96_learn_long_term_depen_narx,siegelmann97_comput_capab_recur_narx_neural_networ}.
For such models, a finite history (or some delays) of $T$ inputs $u_{t-T+1}, \dots, u_t$ is provided to the model at each time step.
Based on this history, the RNN is optimized to predict the future behavior.

Another class of RNNs involve recurrent neural networks with internal states~\cite{elman90_findin_struc_time},
such as gated recurrent units (GRUs) and
long short-term memory (LSTM) neural networks~\cite{hochreiter97_long_short_term_memor,gers00_learn_to_forget},
as well as reservoir computing~\cite{tanaka19_recen_advan_physic_reser_comput}.
In particular LSTM neural networks have gained increasing attention in recent years~\cite{salehinejad17_recen_advan_recur_neural_networ, vlachas18_data_driven_forec_high_dimen},
because they have been observed to surpass other RNN variants on various benchmark tasks~\cite{gers00_recur,gers01_lstm_recur_networ_learn_simpl,gers01_applying_lstm_time_series_pred, greff17_lstm} by coping with the vanishing gradient problem~\cite{bengio94_learn_long_term_depen_with,pascanu12_diffic_train_recur_neural_networ}.
For this class of RNN models, only the current input $u_t$ is provided at each time step.
However, and in contrast to sliding window approaches, the model additionally depends on a set
of internal states: the cell state variables $\vect{c}_t$ and hidden states $\vect{h}_t$.

In other words, sliding window approaches model the unobserved dimensions of the data
using a certain number of delayed observations (``delays'')  as additional input, whereas LSTMs and GRUs model such dimensions using a set of internal states.
While the sliding window is shifted every time step during inference,
LSTMs update their internal states based on certain gating mechanisms.
The usage of a large number of internal states has rendered LSTMs superior
in modeling high-dimensional dynamical systems as they appear, for example, in NLP~\cite{graves13_gener_sequen_with_recur_neural_networ}.

In general, the output of RNNs can either be a prediction of the observed data at a future time step, $\hat{u}_{\tau}$, $\tau>t$ (discrete time dynamics)
or possibly the time derivative of the input at the current time step $\dot{u}_t$; this would give rise to continuous (differential-discrete delay) equations. 

As for any dynamical system, RNNs require an appropriate set of initial conditions to be well defined.
For sliding window approaches, this means that an input sequence observed over some time interval $T$ must be provided for the model to make predictions about the future.
For LSTM-like RNNs, proper initial conditions for the internal states $\vect{c}_0$ and $\vect{h}_0$ are missing,
and the problem is typically dealt with by setting the initial state values either to zero or to random values~\cite{zimmermann12_forecasting_with_rnns}.
To mitigate this inconsistent initialization,
a so-called ``warmup'' or ``washout'' phase is typically employed.
During this phase,
a history of true input data is provided to the RNN,
in the hope that the effect of the initialization has ``decayed'' by the end of the phase.
More recent approaches use the $\vect{c}_0$ and $\vect{h}_0$ as additional
learnable parameters~\cite{becerra02_system_ident_using_dynam_neural_networ},
or learn the initial internal states based on histories of the $u_t$~\cite{mohajerin18_multi_step_predic_dynam_system}.

Here, we propose a systematic alternative initialization approach.
We repeat the argument that the warmup phase can be interpreted as ``driving'' of the LSTM by the true system, a forcing leading to 
effectively synchronizing~\cite{rulkov95_gener_synch_chaos_direc_coupl_chaot_system,kocarev96_gener_synch_predic_equiv_unidir, pecora90_synch_chaot_system}
or slaving~\cite{haken75_gener_ginzb_landau_equat_phase} the learned dynamical system to the true data generating process.
We illustrate this on discrete-time data stemming from the Brusselator,
a simple nonlinear model system exhibiting oscillatory dynamics~\cite{kondepudi14_brusselator_chapter}.
Thereby, only one of the variables is observed and used for training.
We subsequently illustrate how that warmup may require long histories of input data before accurate predictions can be obtained (which may
only be guaranteed as $t \to \infty$).
However, if the time series data is effectively low dimensional,
we show that one can extract an embedding $\manifold$ of the observed data using manifold learning.
Similar to earlier approaches~\cite{sutskever13_on_the_importance,strobelt18_lstmv},
we visualize the dynamics of the internal states $\vect{c}_t$ and $\vect{h}_t$, and
furthermore show that they quickly converge to a low-dimensional (two-dimensional) manifold $\manifoldch$
for the learned model.
On this manifold, we show that we can learn the $\vect{c}_t$ and $\vect{h}_t$ as graphs of a function over the data manifold $\manifold$.
Mapping from short histories of $u_t$ to $\manifold$,
we can thus infer the proper initial states $\vect{c}_0$ and $\vect{h}_0$ for a given input sequence.
This obviates the need for a warmup phase, and, as we will show, leads to accurate predictions.
Our approach is thus similar to the work presented in Ref.~\cite{mohajerin18_multi_step_predic_dynam_system} in that we learn the initial condition based on the input.
However, by learning an intrinsic data manifold, our approach has the advantage of providing a solid theoretical basis (and a minimal input sequence length for cold-starting.)

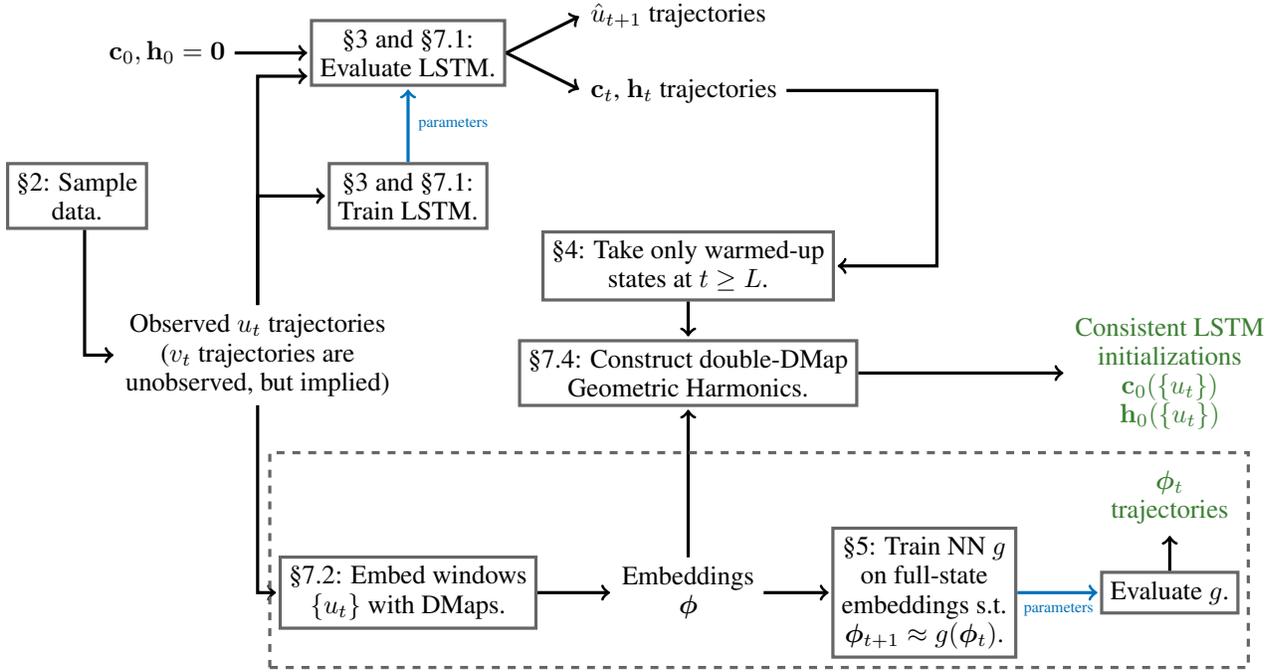
\begin{figure}
    \centering
        \newcommand{\optionalOpacity}{0.65}
\newcommand{\horBoxShift}{2cm}
\newcommand{\vertLineShift}{0.5cm}
\newcommand{\spacingForLSTMIndBox}{2*\vertLineShift}
\newcommand{\figSectionRefs}[1]{#1}

\begin{tikzpicture}[
    expression/.style={rectangle, minimum size=7mm, align=center},
    operation/.style={rectangle, draw=black!60, very thick, minimum size=5mm, align=center},
    myarrow/.style={->, shorten >=1pt, very thick},
    output/.style={OliveGreen},
    train/.style={},
    params/.style={RoyalBlue},
    optional/.style={opacity=\optionalOpacity},
    optionalArrow/.style={dotted,opacity=\optionalOpacity},
]

\node[operation, train]  (trainLSTM) {\figSectionRefs{\S\ref{sec:architecture} and \S\ref{sec:training}:\\}Train LSTM.};
\node[operation]  (evalLSTM)  [above=of trainLSTM] {\figSectionRefs{\S\ref{sec:architecture} and \S\ref{sec:training}:\\}Evaluate LSTM.};
    \draw[myarrow, params] (trainLSTM.north) -- (evalLSTM.south) 
        node[midway,right] {\tiny parameters};

\node[expression] (cZero)     
    [left=of evalLSTM]
    {$\cell_0,\hidden_0=\mathbf{0}$};
    \draw[myarrow] (cZero) -- (evalLSTM);

\node[expression] (observedu) 
    [below=of trainLSTM, xshift=-\horBoxShift] 
    {Observed $u_t$ trajectories\\($v_t$ trajectories are\\unobserved, but implied)};
    \draw[myarrow] (observedu.north) |- (trainLSTM.west);
    \draw[myarrow] (observedu.north)  |- (evalLSTM.base west);
    
\node[operation] (start)
    [above=of observedu,xshift=-1.2*\horBoxShift]
    {\figSectionRefs{\S\ref{sec:data}: }Sample\\data.};

\node[operation] (embedWindows)
    [below=of observedu, xshift=\horBoxShift, yshift=-2*\vertLineShift]
    {\figSectionRefs{\S\ref{sec:dmaps}: }Embed windows\\$\utSeq$ with DMaps.};
    \draw[myarrow] (observedu.south) |- (embedWindows.west);
    
\node[expression] (embeddings)
    [right=of embedWindows]
    {Embeddings\\$\vect\embedding$};
    \draw[myarrow] (embedWindows) -- (embeddings);

\node[operation] (geometricHarmonics)
    [above=of embeddings, yshift=\spacingForLSTMIndBox]
    {\figSectionRefs{\S\ref{sec:GH}: }Construct double-DMap\\Geometric Harmonics.};
    \draw[myarrow] (embeddings) -- (geometricHarmonics);

\node[operation, train] (trainPredictor)
    [right=of embeddings,xshift=-.1cm]
    {\figSectionRefs{\S\ref{sec:state_space}: }Train NN $\predictorNN$\\on full-state\\embeddings s.t.\\
    $\vect \phi_{t+1}\approx g(\vect \phi_t)$.};
    \draw[myarrow] (embeddings) -- (trainPredictor);
    
\node[operation] (evalPredictor)
    [right=of trainPredictor,xshift=.1cm]
    {Evaluate $\predictorNN$.};
    \draw[myarrow, params] (trainPredictor) -- (evalPredictor) 
        node[midway,below] {\tiny parameters};
        optional/.style={opacity=\optionalOpacity},

\node[expression, output] (embeddingTrajectories)
    [above=of evalPredictor,yshift=-\vertLineShift]
    {$\vect\embedding_t$\\trajectories};
    \draw[myarrow] (evalPredictor) -- (embeddingTrajectories);

%

\begin{scope}[node distance=0cm and 0cm]
\node[expression, output] (newInitializations)
    [right=of geometricHarmonics.east -| embeddingTrajectories.north, anchor=center]
    {
        Consistent LSTM\\
        initializations\\
        $\cell_0(\utSeq)$\\
        $\hidden_0(\utSeq)$
    };
\end{scope}
    \draw[myarrow] (geometricHarmonics.east) -- (newInitializations.west);

\draw[myarrow] (start.south) ++(.1, 0) |- (observedu.west);    
    
        
    



\node[operation,dashed] (lstmIndep)
    [fit=
        (embedWindows) (embeddings) 
        (trainPredictor) (evalPredictor) (embeddingTrajectories) 
    ]
    {};

\node[expression] (cthtTraj)  
    [right=of evalLSTM, yshift=-\vertLineShift]              
    {$\cell_t$, $\hidden_t$ trajectories};
    \draw[myarrow] (evalLSTM.east) -- (cthtTraj.west);

\node[expression] (uhatTraj)  
    [right=of evalLSTM, yshift=\vertLineShift]              
    {$\hat{u}_{t+1}$ trajectories};
    \draw[myarrow] (evalLSTM.east) -- (uhatTraj.west);
    
\node[operation] (warmedUp)
    [above=of geometricHarmonics,yshift=-\vertLineShift]
    {\figSectionRefs{\S\ref{sec:manifold_initialization}: }Take only warmed-up\\states at $t\ge \warmedUpThreshold$.};
    \draw[myarrow] (cthtTraj.east) -- ++(2,0) -| ++(0, -.5) |- (warmedUp.east);
    \draw[myarrow] (warmedUp.south) -- (geometricHarmonics.north);

\end{tikzpicture}
    \caption[Approach to initialize LSTM neural networks using manifold learning and interpolation.]{
        \textbf{Schematic initialization approach for LSTM neural networks using manifold learning and interpolation.}
        Typically LSTM are trained to model trajectories of an observed variable $u$, 
        using a naive all-zeros initialization for the hidden states $\hidden_0$ and cell states $\cell_0$.
        The trained LSTM can be run in autoregressive mode to produce time-series estimates $\hat{u}_{t+1}$
        for the observed variable (and, as a byproduct, time series of the internal states $\hidden_t$ and $\cell_t$) 
        using a sufficiently long warmup period.
        Note that our indices for the LSTM inputs $u_t$,
        $\vect{c}_{t-1}$, and $\vect{h}_{t-1}$
        differ by one in deference to the established LSTM literature.
        
        Simultaneously, we use diffusion maps, 
        an unsupervised manifold learning technique, 
        to embed windows $\utSeq$ of observations.
        This is marked by a dashed border around the lower part of the figure 
        and is independent of
        the LSTM training.
        In other words, we learn the manifold $\manifold$
        on which the data lives using manifold learning.
        For an LSTM neural network trained on the observed variable $u$ with internal states
        $\vect{c}_t$ and $\vect{h}_t$, one finds that the dynamics of each of the internal states 
        quickly converge to (are attracted to) a two-dimensional manifold 
        parametrized by the data-driven coordinates $\manifoldch_c$ and $\manifoldch_h$, even without forcing.
        
        Using geometric harmonics, one can construct the function from $\manifold$ to $\manifoldch_{c,h}$, 
        and thus to the \textit{mature} internal states $\vect{c}_t$ and $\vect{h}_t$.
        This means that one can find the consistent initial states $\vect{c}_0$ and $\vect{h}_0$ 
        for an initial, sufficiently long window of observed data $u_t$ 
        via mapping to the data manifold $\manifold$.
        Alternatively, one can also learn the full state space model $\predictorNN(\phi^{(1)}, \phi^{(2)})$ %
        on the data manifold $\manifold$ (see \sect~\ref{sec:state_space}).
        
        Finally, we note that, since the embeddings $\vect\phi$ are posited to contain information
        for $v$ as well as $u$, we can also devise an imputation method to densely fill in sporadically available $v$ data (see \sect~\ref{sec:discussion}).
        %
    }
    \label{fig:work_flow}
\end{figure}

The steps described above are summarized in \fig~\ref{fig:work_flow}.
The data generating process creates time series of observed variables $u_t$ and unobserved variables $v_t$.
Based on the observed sequential data $u_t$, one can train an LSTM model mimicking the data generating process.
However, as discussed, proper initial $u_0$, $\vect{c}_0$ and $\vect{h}_0$ values are
needed for prediction.
Here, we propose to learn the data manifold $\manifold$ using diffusion maps, a nonlinear manifold learning technique~\cite{Coifman2006}.
As we will show, one can learn the internal states as functions on this manifold, enabling the inference of proper initial states $\vect{c}_0$ and $\vect{h}_0$
for a given minimal $u_t$ input sequence. Here, we use geometric harmonics (GH) to learn this mapping.
Finally, having the data manifold $\manifold$ also facilitates, if desirable and given sufficient data, learning the full-state dynamical system $g$ directly.

The remainder of this article is organized as follows.
In \sect~\ref{sec:data}, we introduce the illustrative model system,
and describe how the data under study is created.
The LSTM model architecture is described in more detail in \sect~\ref{sec:architecture}.
In this section, we also reiterate
how the warmup phase can be viewed as a (generalized) synchronization of the learned dynamical system with the true data generating process.
In \sect~\ref{sec:manifold_initialization}, we describe in more detail
how we extract the data manifold using manifold learning.
Based on the dimension of this manifold, we discuss the minimal number of time steps needed for a proper initialization of the LSTM neural network.
We subsequently evaluate 
``proper'' or ``mature''  initial states $\vect{c}_0$ and $\vect{h}_0$
as functions on this data manifold, and outline the advantages of the approach.
The data manifold also provides a scaffolding on which to learn a full state space model, as discussed
in \sect~\ref{sec:state_space}; we conclude with a discussion of the presented method and open questions in \sect~\ref{sec:discussion}.
Details on the methods used are summarized in the Methods section~\ref{sec:methods}.

\section{A simple illustration: the oscillatory Brusselator}
\label{sec:data}

\begin{figure}
    \centering
    \includegraphics[width=\textwidth]{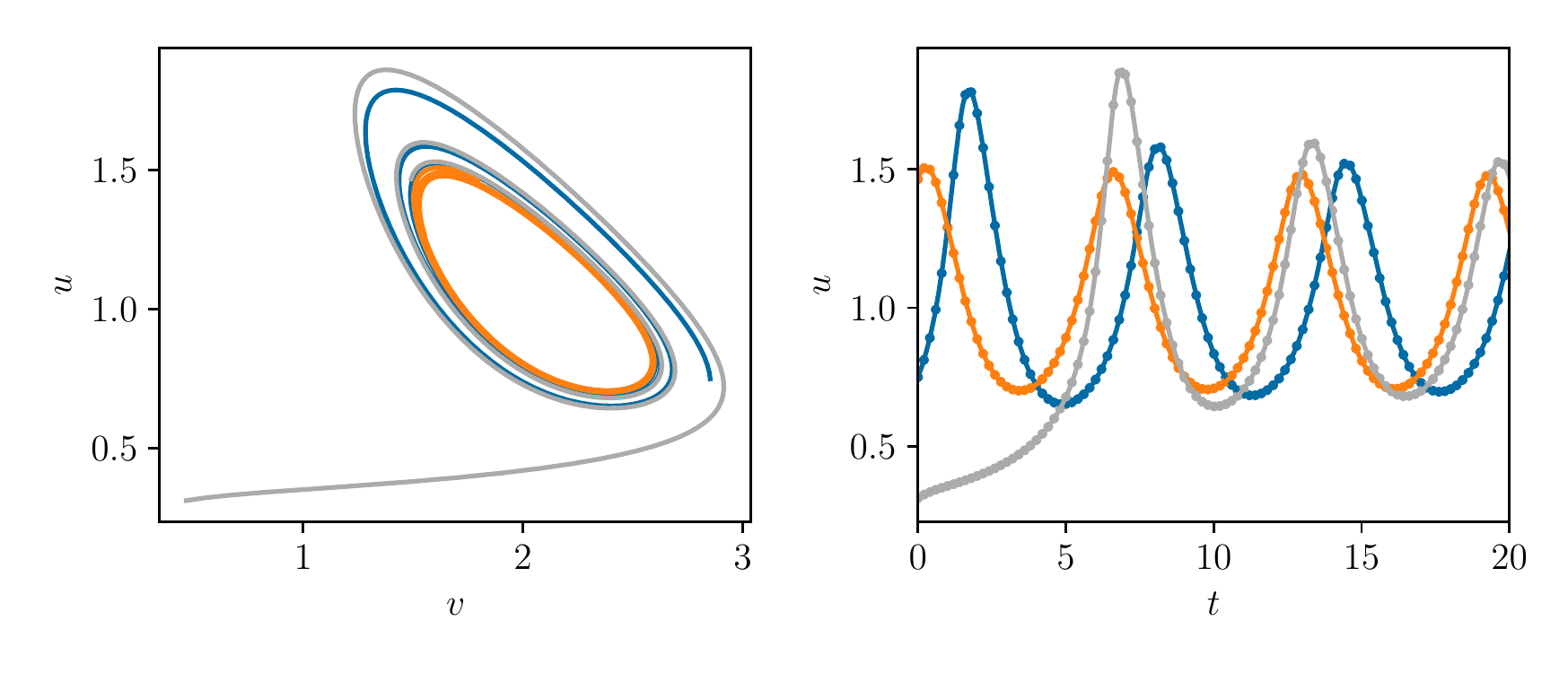}
    \caption{Dynamics of the Brusselator for $a=1$ and $b=2.1$.
      As shown (through color) for three different initial conditions,
      the trajectories of the continuous-time dynamical system settle onto a limit cycle (left).
      Here, we sample the $u(t)$ time series of such trajectories at discrete steps in
      time $\delta t = 0.2$ apart, as illustrated on the right.}
    \label{fig:data}
\end{figure}

In the following sections, we use the Brusselator as a model nonlinear dynamical
system~\cite{kondepudi14_brusselator_chapter}.
This is a prototypical chemical kinetic scheme exhibiting oscillatory instabilities.
It is given by
\begin{align}
    \dot{u} &= a + u^2v - (b+1)u,\label{eq_:1a}\tag{1a}\\
    \dot{v} &= bu - u^2v,\label{eq_:1b}\tag{1b}
\end{align}
\setcounter{equation}{1}\noindent
with two real variables $u$ and $v$ and two real parameters $a$ and $b$, which we keep fixed at $a=1$ and $b=2.1$.
(Note here, and we will revisit this in the discussion, that our approach also works for problems that are parameter dependent.)
The variables $u$ and $v$ can be viewed as re-scaled concentrations of two chemical species.
For this set of parameters $a$ and $b$,
the only attractor of the Brusselator is a stable limit cycle,
see \fig~\ref{fig:data} (left) for three trajectories starting from different initial conditions in
phase space.
Here, we assume that $v$ is unobserved, and we sample $u$ at equidistant points in time.
In particular, we take $N=500$ different initial conditions, and integrate the Brusselator for an interval of $20$ dimensionless time units from each. We then sample $u$ every $\delta t = 0.2$ time steps.
Three sample $u$ time series are depicted in \fig~\ref{fig:data}~(right).
In the following, we denote $u_t$ the discrete time trajectories obtained
from the continuous-time Brusselator, as described above.

\section{LSTM-Model for Learning the Discrete-Time Brusselator Dynamics}
\label{sec:architecture}

Using the set of observed trajectories $u_t$, we set out to learn temporal evolution using a
recurrent neural network (RNN) with internal states~\cite{elman90_findin_struc_time}.
Many different such RNN architectures exist that one may employ for such a task, such as gated recurrent units (GRUs)~\cite{cho14_learning} or long short-term memory (LSTM) cells~\cite{hochreiter97_long_short_term_memor,gers00_learn_to_forget}. 
See also Ref.~\cite{salehinejad17_recen_advan_recur_neural_networ} for an overview of RNN variants.
It is worth mentioning that for low-dimensional prediction problems, such as the dynamics of the Brusselator system,
sliding-window approaches offer a promising alternative~\cite{gers01_applying_lstm_time_series_pred}. 
%

\begin{figure}
    \centering
    \includegraphics{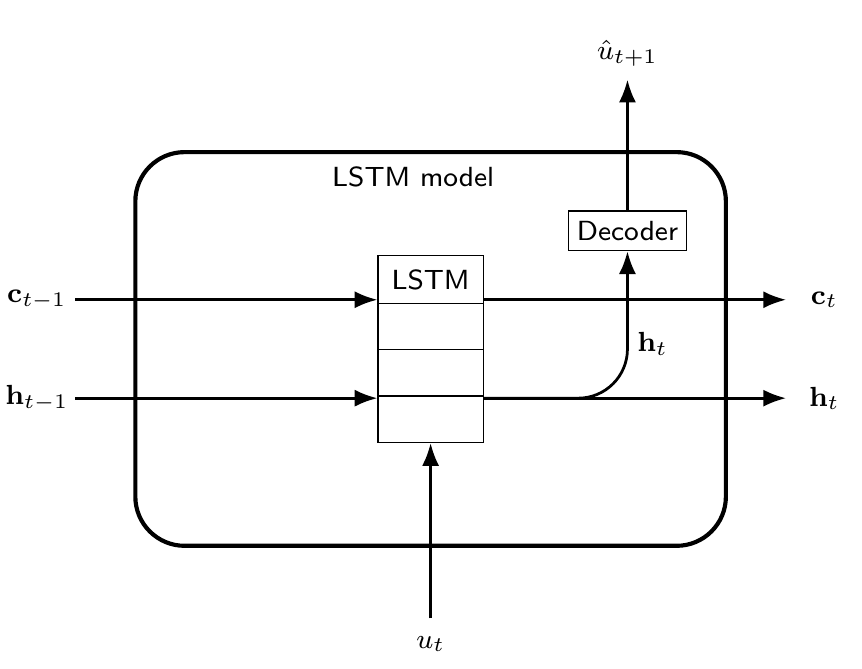}
    \caption{LSTM model for the prediction of the observed variable $u_{t+1}$ of the Brusselator system.
      At each time step, the observed variable $u_{t}$ is fed into the four LSTM cells.
      The cell states $\vect{c}_{t-1}$ and the state vector $\vect{h}_{t-1}$ are then progressed forward in time. The resulting hidden state vector $\vect{h}_{t}$ is decoded using a linear layer, yielding the prediction $\hat{u}_{t+1}$. Both $\vect{c}_t$ and $\vect{h}_{t}$ are then passed directly to the next time step. When used in an autoregressive manner, the predictions $\hat{u}_{t+1}$ are used as input for the next time period.
    }
    \label{fig:architecture}
\end{figure}

However, through the Markov property of gated RNNs, such as LSTM networks,
such models, when correctly initialized, can be used in an autoregressive fashion, using information from only the current time
step. This is opposed to sliding-window approaches such as multilayer perceptrons~\cite{HUDSON19902075},
time-delay neural networks and NARX networks~\cite{gers01_applying_lstm_time_series_pred},
which require an input sequence for prediction.
Here, we use a single-layer LSTM model with $D=4$ LSTM cells, that is with a hidden state vector $\vect{h}_t\in \mathbb{R}^D$, cell states $\vect{c}_t\in \mathbb{R}^D$ and one-dimensional input $u_t$.
At each time step $t$, we progress $\vect{h}_{t-1}$ and $\vect{c}_{t-1}$ forward in time
using the input $u_t$, returning $\vect{h}_{t}$ and $\vect{c}_{t}$.
We subsequently map the $\vect{h}_{t}$ back to the observed variable $\hat{u}_{t+1}$ using
a linear transformation (decoder),
\begin{equation}
  \hat{u}_{t+1} = \mat{W}\vect{h}_{t} + b.
\end{equation}
The model is then optimized using the mean-squared error between the predicted next time step values $\hat{u}_{t+1}$ and their true observations $u_{t+1}$ based on input sequences
using backpropagation through time~\cite{bptt1,bptt2,bptt3}.
See \fig~\ref{fig:architecture} for details and a schematic of the model and \sect~\ref{sec:training} for details on the training.
Note that, for parameter-dependent problems, the parameter can simply be appended as an additional input, along with $u_t$ at each time step.

An issue arises when trying to model internal variables as in the case discussed here.
In particular, an initial condition must be specified for the cell states $\vect{c}_t$ and hidden states $\vect{h}_t$ at $t=0$.
This means, given the observed variable $u_1$, we need the internal vectors $\vect{c}_0$ and $\vect{h}_0$ in order to predict the next states $\vect{c}_1$, $\vect{h}_1$.
For training RNNs, as for example neural networks based on LSTM cells, earlier approaches initialized $\vect{c}_0$ and $\vect{h}_0$ with just zeros,
or random values~\cite{zimmermann12_forecasting_with_rnns}.
This approach, however, requires a washout (also called warmup phase, synchronization phase or listening time)
by providing a longer history of $u_t$ values to the model~\cite{mohajerin18_multi_step_predic_dynam_system}.
The idea is that the effect of the inconsistent initialization of the internal states (inconsistent with the observed history of $u_t$ values) will decay after a sufficiently long washout phase;
and that when the model is used in an autoregressive fashion after this initial phase, the predictions remain accurate and consistent with past observations.

\begin{figure}
    \centering
    \includegraphics[width=\textwidth]{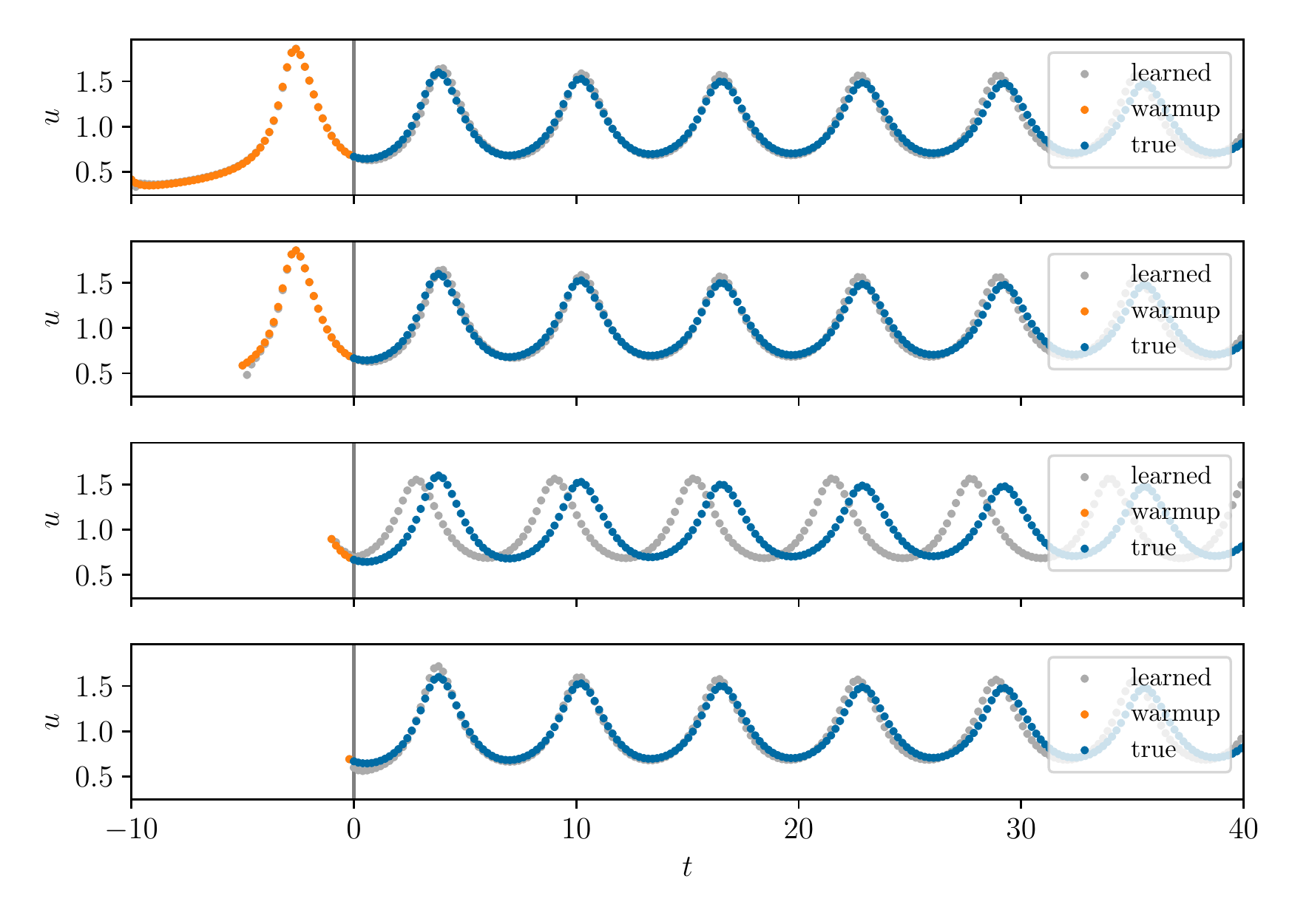}
    \caption{Predictions using the learned LSTM model for different amount of warmup
      steps $T_{\mathrm{warmup}}$.
      From top to bottom, $T_{\mathrm{warmup}}=50, 25, 5$,
      whereas in the last row only a single $u_1$ value
      is presented to the model, and thus no warmup is used.
      The warmup steps are indicated in orange, with blue marking the true $u_t$ trajectory and
      gray representing the values obtained from prediction.
      With a reduced amount of warmup, the influence of the ``wrong'' initial condition $\vect{c}_0=0$ and $\vect{h}_0=0$ becomes stronger and thus the accuracy worse. However, by chance (last row), the dynamics may still converge to the right phase on the limit cycle.}
    \label{fig:integration}
\end{figure}

This implies that the ``mature'', long-term behavior of the LSTM lies on a two-dimensional invariant manifold (corresponding to the Brusselator dynamics); and that this ``Brusselator manifold'' is {\em attracting}
in the higher dimensional LSTM state space.
See \fig~\ref{fig:integration} for predictions of the learned LSTM model for different washout periods.
In particular, one can observe that too short of a warmup phase can lead to wrong phase predictions
when the model is used autoregressively (the long-term attractor is still correct, but the phase on it will, in general, be wrong). 
More recent approaches use the $\vect{c}_0$ and $\vect{h}_0$ as additional learnable parameters~\cite{becerra02_system_ident_using_dynam_neural_networ},
or learn the $\vect{c}_0$ based on histories of the $u_t$~\cite{mohajerin18_multi_step_predic_dynam_system}.

We repeat the well-known argument that, in fact, one can view the washout phase as the synchronization of two dynamical systems. One system, the learned RNN,
is driven by one variable of the true data generating process~\cite{pecora90_synch_chaot_system, strogatz14_book}.
If/when these two dynamical systems synchronize, then the driven system becomes \textit{slaved} to the dynamics of the forcing system~\cite{haken75_gener_ginzb_landau_equat_phase}.
In the following, we treat the RNN, \textit{after} successfully optimizing its parameters, as a surrogate model for the Brusselator and investigate its "entrainment" by the continuous-time problem.

Assume that $u(t)$ is a time series of the true Brusselator system.
Furthermore, assume that we have a (continuous-time) NN model that has been successfully trained to represent the Brusselator dynamics with variables $u_{\text{nn}}$ and $v_{\text{nn}}$.
Then the dynamics of $v_{\text{nn}}$ follows
\begin{equation}
\dot{v}_{\text{nn}} = bu_{\text{nn}} - u_{\text{nn}}^2 v_{\text{nn}},
\end{equation}
as for the original Brusselator system Eqs.~\eqref{eq_:1a}-\eqref{eq_:1b}.
We can now interpret initialization through warmup as a forcing of the learned hidden Brusselator variable $v_{\text{nn}}$ by the true time series data $u(t)$. This means that the dynamics of $v_{\text{nn}}$ follow
\begin{equation}
\dot{v}_{\text{nn}} = bu(t) - u(t)^2 v_{\text{nn}}.
\end{equation}
One can now define the synchronization error of the true Brusselator variable $v$ and the learned variable
$v_{\text{nn}}$ as $e_1 = v-v_{\text{nn}}$, yielding the linear dynamics
\begin{equation}
\dot{e}_1 = - u(t)^2 e_1.
\end{equation}
Since $u(t)^2\geq 0$, this means the synchronization error decays to zero for $t\rightarrow \infty$, and the learned model synchronizes with the true dynamics of the Brusselator.

In general, the state dimension of the RNN model will be different from the dimension of the true underlying dynamical system (as in the LSTM model discussed above, where the cell state dimension is larger than the dimension of $v$).
However, analogous to the case discussed in this section, the theory of \textit{generalized synchronization}~\cite{rulkov95_gener_synch_chaos_direc_coupl_chaot_system,kocarev96_gener_synch_predic_equiv_unidir} states that the cell states will asymptotically become a function of the forcing~\cite{pecora90_synch_chaot_system}.
This property of generalized synchronization can be proven for some classes of echo state networks in reservoir computing~\cite{jaeger01_the_echo_state,jaeger04_harnes_nonlin,tanaka19_recen_advan_physic_reser_comput}.
In such echo state networks, the assumption is that the influence of the initial condition
has vanished after a washout phase, which for such models is called the \textit{echo state property}~\cite{jaeger01_the_echo_state,yildiz12_re_visit_echo_state_proper,lu18_attrac_recon_by_machin_learn}.
For general LSTM neural networks, such a proof is still missing.

\section{Initialization through Manifold Learning}
\label{sec:manifold_initialization}

\begin{figure}
    \centering
    \includegraphics[width=0.5\textwidth]{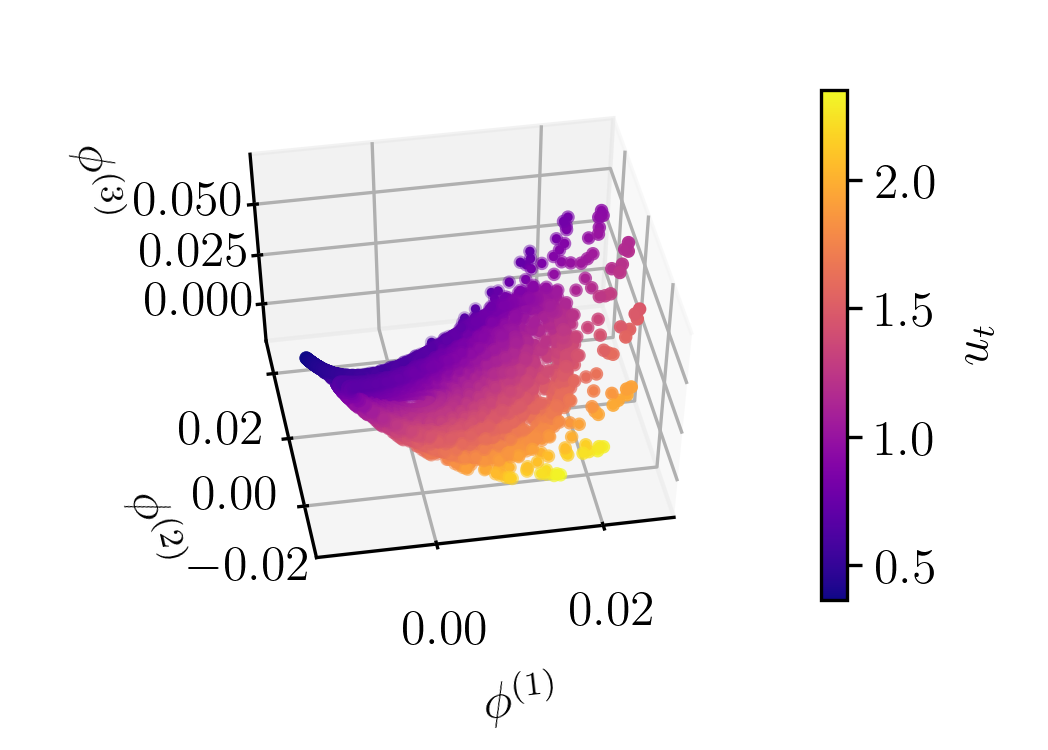}
    \caption{Data manifold obtained using diffusion maps on time series windows $u_t$ of length $l=10$,
      indicating that the data is in fact two-dimensional.
      The color encodes the initial value $u_1$ corresponding to each window.}
    \label{fig:dmaps_on_input}
\end{figure}

Initialization using a washout period, that is, through synchronization as discussed above,
has the drawback that complete synchronization can only be expected after an
infinitely long washout phase.
However, as we will show in this section, when the dynamics are low-dimensional,
one can in fact infer consistent initial conditions $\vect{c}_0$ and $\vect{h}_0$ from short histories of observed data $u_t$.
We do this by learning the low-dimensional manifold on which the
observed $u_t$ trajectories live.
In order to do so, we sample time series windows comprising $l=10\geq 2n+1$ discrete time steps\footnote{As we will discuss later, the exact length is not important, as long as $l \geq 2n+1$, with $n$ being the intrinsic dimension of the system manifold.}.
On this collection of short time series, we perform manifold learning in order to find a lower-dimensional embedding.
Here, our method of choice is diffusion maps, a nonlinear manifold learning technique.
See \sect~\ref{sec:dmaps} for a detailed description of the method.
Using diffusion maps, we find that the collection of $u_t$ time series live on a two-dimensional manifold spanned by the two leading diffusion modes $\phi^{(1)}$ and $\phi^{(2)}$, cf. \fig~\ref{fig:dmaps_on_input}.
This finding is in agreement with the two-dimensional data generating process, Eqs.~\eqref{eq_:1a} and~\eqref{eq_:1b}.
Furthermore, this means that every $u_t$ trajectory window is uniquely defined by a small set of generic observations; in principle one needs $2n+1=5$ such observation, but here we were lucky, and just two of our diffusion map coordinates sufficed.
For the Brusselator, if both variables could be measured, $u(t=0)$ and $v(t=0)$ would suffice. 
In our case, short histories of $u_t$ serve as the sufficiently rich observations.

\begin{figure}
    \centering
    \includegraphics[width=\textwidth]{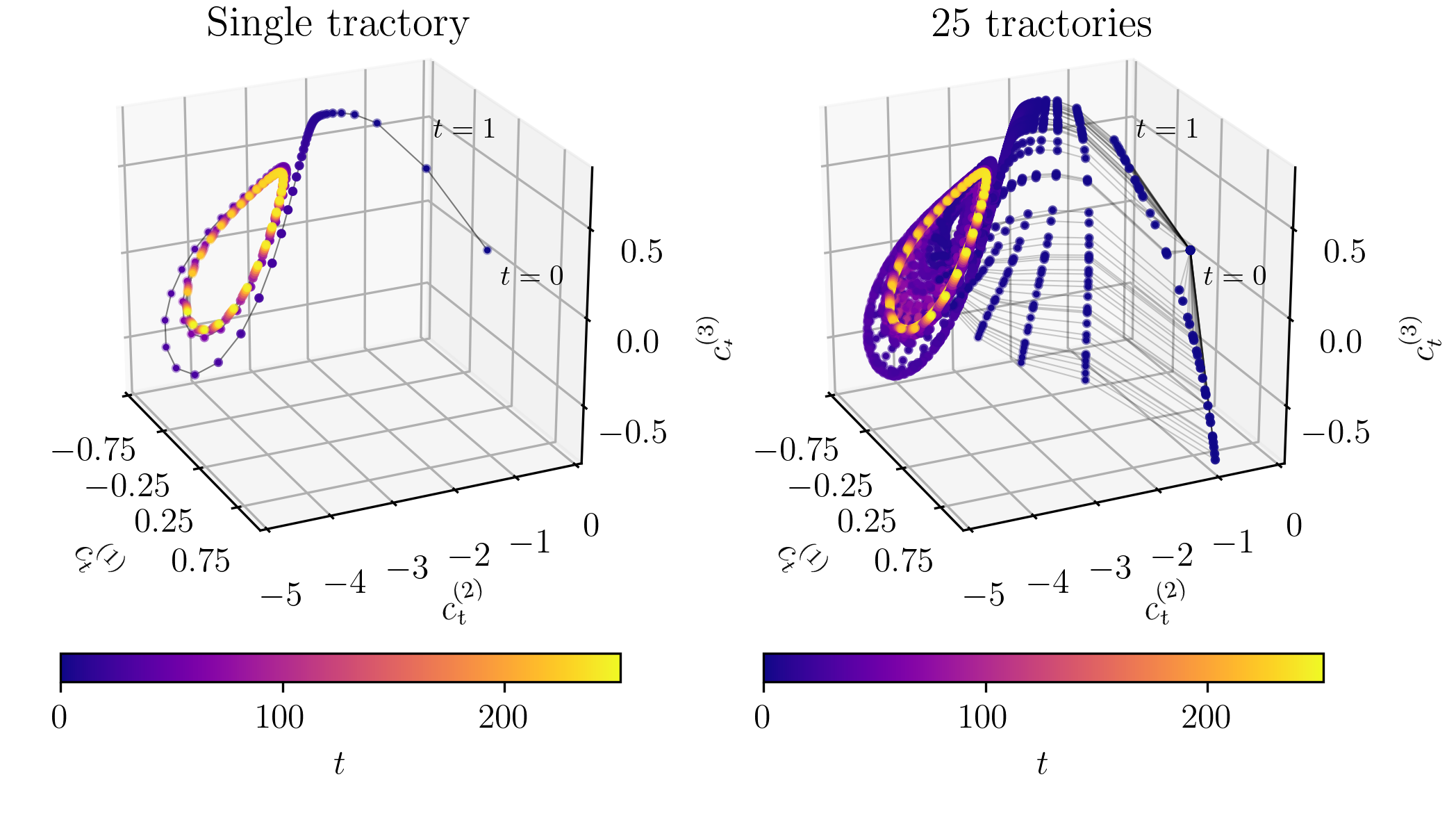}
    \caption{The hidden cell states $\vect{c}_t$ obtained by iteration providing an
      initial $u_1$ value but no further warmup, projected onto the first three variables $c_t^{(1)}$, $c_t^{(2)}$ and $c_t^{(3)}$. On the left, the projection of the $\vect{c}_t$ values of only a single trajectory is shown, whereas on the right the values of 25 trajectories are depicted. The color corresponds to the time step $t$. As obvious from the color coding, the dynamics settles onto a limit cycle for large $t$, as expected. Furthermore, the $\vect{c}_t$ quickly come down to a two-dimensional manifold.
      For better visibility, the trajectories of the $\vect{c}_t$ values are indicated by thin black lines. See \sect~\ref{sec:methods} for a projection onto $c_t^{(1)}$, $c_t^{(2)}$ and $c_t^{(4)}$, for completeness.}
    \label{fig:dmaps_ct_wo_warmup}
\end{figure}

In a similar way, we can also look at projections of the cell state vectors $\vect{c}_t$ of the trained LSTM model.
The projections onto $c_t^{(1)}$, $c_t^{(2)}$ and $c_t^{(3)}$ for a single trajectory and for a collection
of 25 trajectories, starting from different initial conditions, are depicted in \fig~\ref{fig:dmaps_ct_wo_warmup} on the left and right, respectively.
Here, the trajectories are colored with the discrete time $t$.
See \sect~\ref{sec:GH} for a projection of the cell states onto $c_t^{(1)}$, $c_t^{(2)}$ and $c_t^{(4)}$ and projections of the hidden states $\vect{h}_t$.
Note that since the $\vect{c}_t$ are initialized at 0, all trajectories initially start at the same point.
In addition, an initial condition $u_1$ is provided to the model, but no further warmup phase is provided, and the model is used autoregressively for prediction.
This initial $u_1$ value results in a one-dimensional spread out of the
cell states $\vect{c}_1$ at $t=1$,
yielding a one-dimensional curve in \fig~\ref{fig:dmaps_ct_wo_warmup}~(right).
After a few iterations, the cell states have converged onto a two-dimensional manifold,
and will eventually converge to a limit cycle in this manifold, as obvious through the coloring.
The same holds for the hidden state $\vect{h}_t$ trajectories, see \sect~\ref{sec:GH} for the complementary projections onto $c_t^{(1)}$, $c_t^{(2)}$ and $c_t^{(4)}$ and of the hidden states $\vect{h}_t$

A few points are worth noting here:
\begin{itemize}
\item First, notice the fast convergence of the $\vect{c}_t$ onto a two-dimensional manifold
  even without warmup.
  This indicates that the model has learned to quickly forget the un-physical initialization $\vect{c}_0=0$, and that the \textit{physical} two-dimensional manifold is not only invariant but also (strongly) attracting.
  After some number of timesteps $\warmedUpThreshold$,
  the $\vect{c}_t$ and $\vect{h}_t$
  become slaved to the corresponding $u_t$ and unobserved $v_t$,
  which fact we will exploit later for initialization.
  This fast convergence must, of course, also depend on the details of the training algorithm. For the record, our illustrative LSTM was trained with a loss computed over the entire available time domain, in contrast to the occasional practice of not penalizing the first few predictions.
  We believe that this leads to accelerated convergence of the internal states to the physical manifold.
\item Intuitively, one could expect that the $\vect{h}_t$ represent the observed variable $u_t$ (since the output of the LSTM is only based on $\vect{h}_t$) and that the $\vect{c}_t$ encode the unobserved variable $v_t$. 
However, we find instead that the cell states and the hidden states
both separately encode a two-dimensional system.
  We conjecture that the LSTM gating structure and the off-manifold initialization of the
  cell states may force the model into learning a combination of the $u_t$ and $v_t$ dynamics
  in the cell states as well as in the hidden states.
\item Finally all trajectories converge to the limit cycle after many iterations,
  in agreement with the observed dynamics of the Brusselator.
  However, {\em on which phase} on the limit cycle the dynamics converges strongly depends on the length of
  the warmup period provided to the model.
  This issue can already be observed from the trajectories shown in \fig~\ref{fig:integration},
  where the integration may converge to the wrong phase if not enough warmup has been provided.
  Only if after warmup has been provided the dynamics of the LSTM model
  predicts the true $u_t$ values at each time step, and thus converges to the right phase on the limit cycle.
\end{itemize}

\begin{figure}
    \centering
    \includegraphics[width=\textwidth]{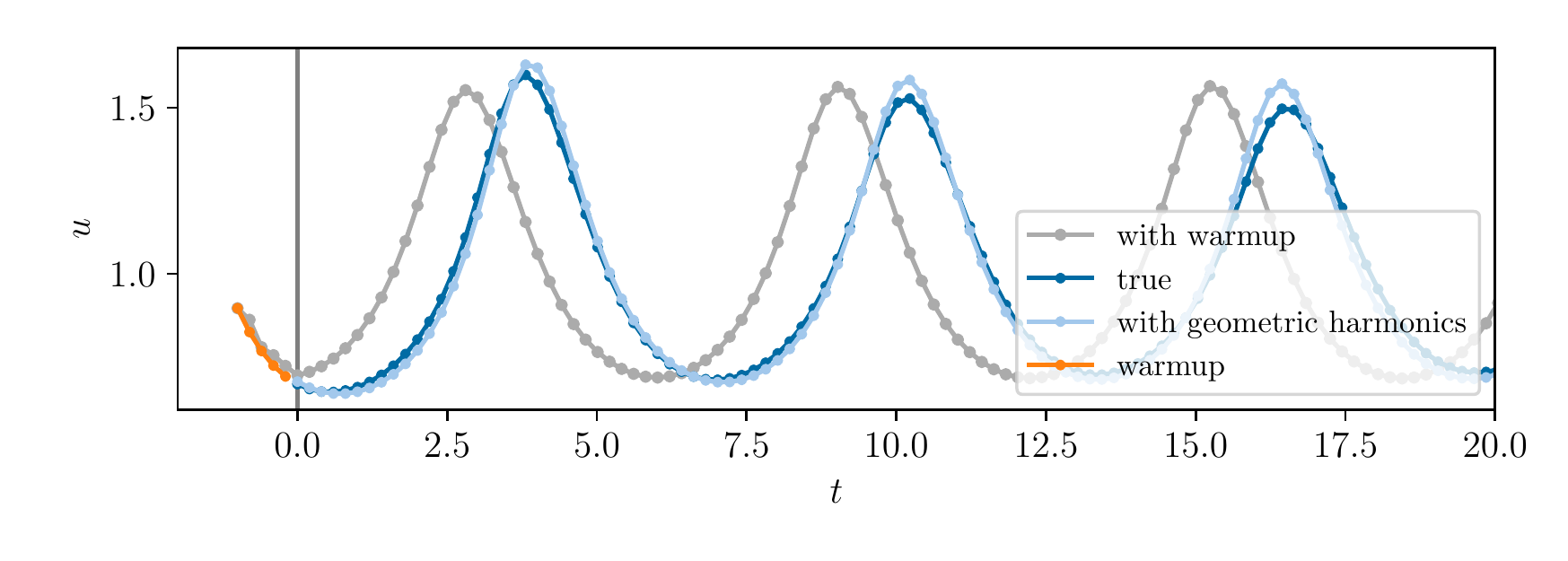}
    \caption{Prediction results using warmup (gray) and using initialization that employs geometric harmonics (light blue). The five steps long warmup phase is indicated in orange, which also corresponds to the $u_t$ chunk used to get $\vect{c}_0$ and $\vect{h}_0$ using Nystr\"{o}m extension and geometric harmonics. Note that although the same amount of data is used for initialization/warmup, the trajectory obtained using warmup converges to wrong phase on the limit cycle.}
    \label{fig:integration_w_gh}
\end{figure}

\begin{figure}
    \centering
    \includegraphics[width=\textwidth]{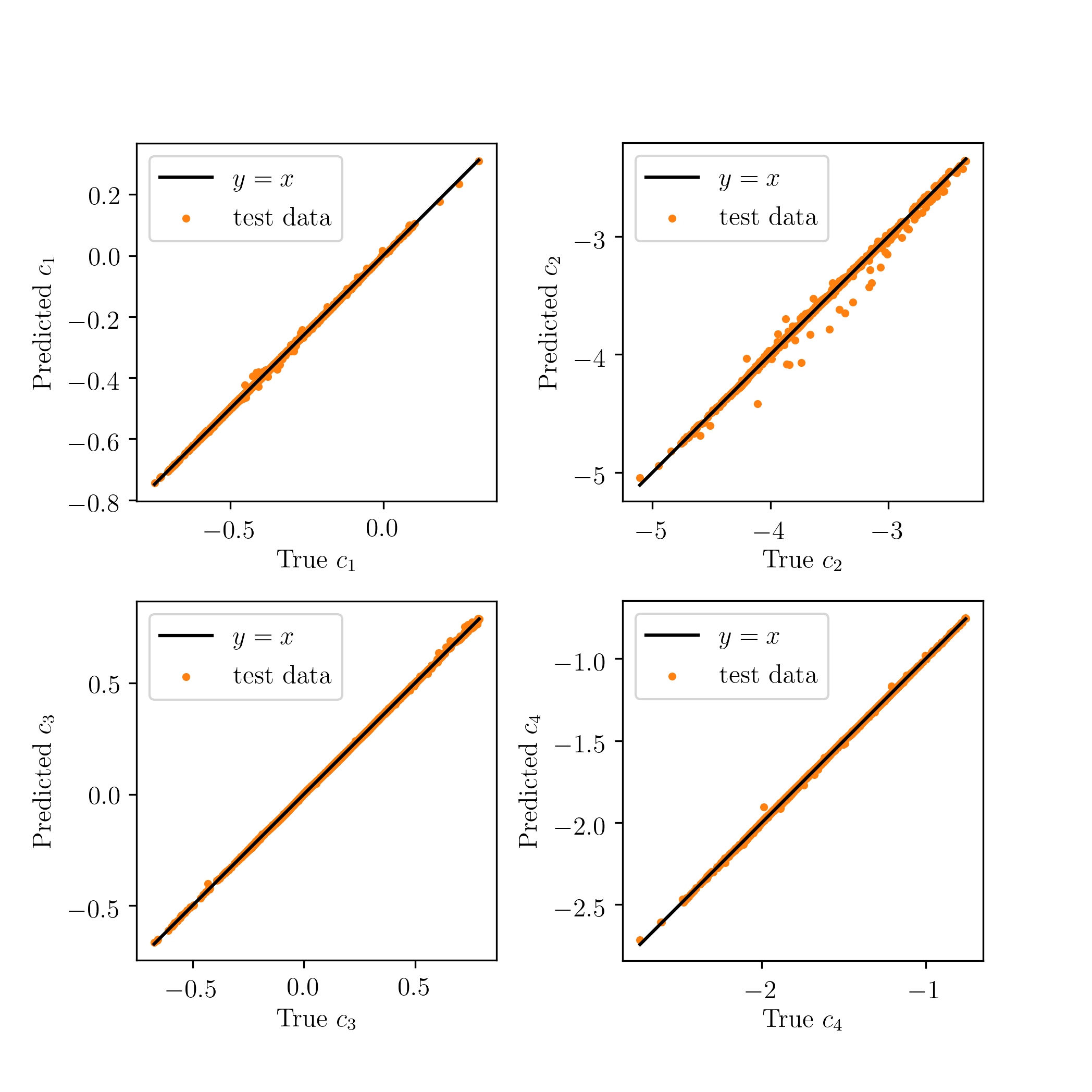}
    \caption{Interpolation of the $\vect{c}_t$ values based on the $\manifold$ embedding and using geometric harmonics.}
    \label{fig:geometric_harmonics}
\end{figure}

Since the $\vect{c}_t$ and $\vect{h}_t$ converge to a two-dimensional manifold after the effects of the un-physical initialization have decayed, we can also infer the consistent $\vect{c}_0$ and $\vect{h}_0$ on this low-dimensional manifold for a given $u_t$ time window.
Assuming we have reached generalized synchronization,
that is, that the internal states have converged to the two-dimensional manifold 
and have become a function of the input data,
we can learn the mapping from the data manifold to the internal state vectors.
We do this by using geometric harmonics (see \sect~\ref{sec:GH}), mapping from the input data manifold
($\phi^{(1)}$, $\phi^{(2)}$) to the $\vect{c}_t$ and $\vect{h}_t$.
Effectively, we skip ahead to the point at which $\vect{c}_t$ and $\vect{h}_t$ are slaved to the given $u_t$ sequence.
Crucially, we construct this mapping
using only $\vect{c}_t$ for $t>\warmedUpThreshold=10$,
where we select the threshold $\warmedUpThreshold$
such that the synchronization has been reached.
%
See \fig~\ref{fig:geometric_harmonics} for predictions of this mapping on test input data after learning.

Having obtained this mapping means that we can go from a short input sequence $u_t$ (here of length 5)
to the data manifold
($\phi^{(1)}$, $\phi^{(2)}$), and subsequently to the internal states.
This means that, for an input sequence $u_t$ we can obtain consistent $\vect{c}_0$ and
$\vect{h}_0$ state vectors.
For $u_t$ trajectories not contained in the training data, the mapping $u_t \mapsto \manifold$
can be extended using Nystr\"{o}m extension, cf. \sect~\ref{sec:Nystrom Extension}.
The dimension of the data manifold thereby also indicates how long the input sequence must
be to obtain a unique corresponding ($\phi^{(1)}$,$\phi^{(2)}$) value and thus a proper
initial cell state vector $\vect{c}_0$.
For the two-dimensional data manifold $\manifold$ obtained above,
and using the Takens embedding theorem~\cite{Takens1981}, $2n+1=5$ observations are prescribed,
and thus an $l=5$ long $u_t$ sequence, is sufficient.
It is worth noting here that, due to the discrete time nature of the dynamical systems, preimages in time might not be unique,
and thus there might be various consistent $\vect{c}_0$ and $\vect{h}_0$ vectors
for a given input $u_t$ sequence~\cite{gicquel98_noninvertability,rico-martinez92_discr_vs}.
The approach discussed above has the advantage that the performance of the trained LSTM model, when used for prediction, does not rely on the washout of the initial condition, and, as we argue, will thus be more accurate. This is also illustrated in \fig~\ref{fig:integration_w_gh},
where a five time step long input sequence is integrated forward (a) using the learned LSTM model employing warmup (gray) and (b) using the initial internal states inferred to correspond to the input sequence (light blue).
It is visually clear that there is a much better agreement between the prediction results from the
consistent initialization using geometric harmonics,
as opposed to the trajectory obtained using a warmup phase.
In particular, the trajectory obtained using warmup converges to the wrong phase on the limit cycle,
indicating that the warmup phase was too short for the model to converge to the right position on the attracting, invariant two-dimensional manifold corresponding to the true Brusselator.

\section{A Next Step: Learning a State Space Model on the Data Manifold}
\label{sec:state_space}

Having learned the data manifold $\manifold$, cf. \fig~\ref{fig:dmaps_on_input}, one can learn the
dynamics of the two independent diffusion components $\vect{\phi}^{(1)}$ and $\vect{\phi}^{(2)}$ instead of training an LSTM; alternatively, we could also learn a continuous-time version of the model.
This means we can transform the task of learning the dynamics of some observed variables $u$ and some unobserved variables $v$ into a problem where we have only observed variables.
In the following, we learn the function $g$ such that
\begin{equation}
    \phi^{(1)}_{t+1}
    ,
    \phi^{(2)}_{t+1}
    = \predictorNN\left(\phi^{(1)}_{t}, \phi^{(2)}_{t}\right).
\end{equation}

\begin{figure}
    \centering
    \includegraphics[width=\textwidth]{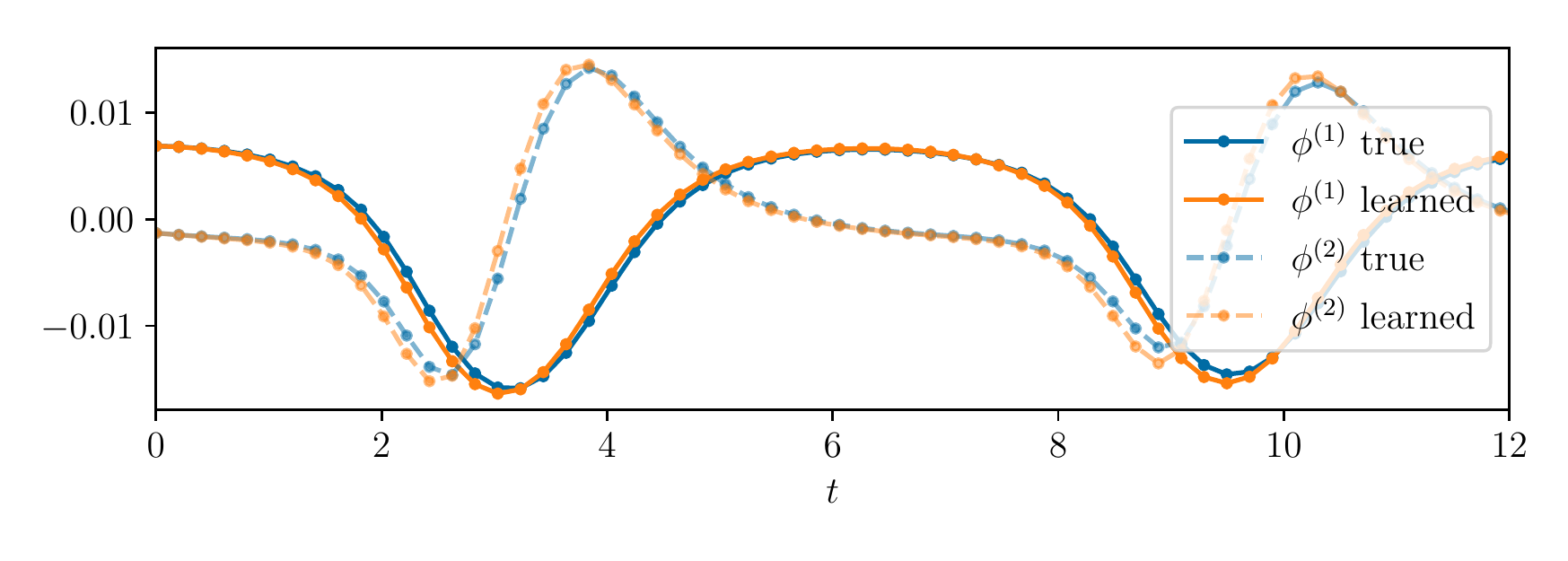}
    \caption{Prediction results of the learned dynamical system $\predictorNN\left(\phi^{(1)}_{t}, \phi^{(2)}_{t}\right)$ (orange) together with the true trajectories (blue) of an initial condition from the test set.}
    \label{fig:integration_phi}
\end{figure}
Here, we represent $g$ through a fully connected neural network with three hidden layers with $64$ neurons each, each hidden layer followed by a Swish activation function~\cite{ramachandran17_swish}.
The model $g$ is optimized using pairs of consecutive $\phi^{(1)}_{t}$, $\phi^{(2)}_{t}$ tuples, each tuple corresponding to a five time steps long $u_t$ trajectory of observed variables.
That is, we use the $\phi^{(1)}_{t}$, $\phi^{(2)}_{t}$ values of two $u_t$ trajectories shifted by just one time step, yielding the pairs of tuples $\left(\phi^{(1)}_{t}, \phi^{(2)}_{t}\right)$, $\left(\phi^{(1)}_{t+1}, \phi^{(2)}_{t+1}\right)$.
The model is then optimized by minimizing the mean squared error between its predictions $\left(\hat{\phi}^{(1)}_{t+1}, \hat{\phi}^{(2)}_{t+1}\right)$ and the true $\left(\phi^{(1)}_{t+1}, \phi^{(2)}_{t+1}\right)$ using the Adam optimizer. Again, a small subset of $u_t$ trajectories (and their corresponding $\left(\phi^{(1)}_{t}, \phi^{(2)}_{t}\right)$ values) are held out as a validation set to assess whether the model has overfitted.
After training, we use the learned dynamical system $g$ to predict an initial condition
$\left(\phi^{(1)}_{0}, \phi^{(2)}_{0}\right)$ which the model has not seen before.
The resulting $\left(\phi^{(1)}_t,\phi^{(2)}_t\right)$ trajectories produced by $g$, together with the true trajectories obtained from the true $u_t$ trajectory, are shown in \fig~\ref{fig:integration_phi}.

\section{Discussion}
\label{sec:discussion}

In this work we proposed a way of finding a data-consistent initialization of trained LSTM neural networks.
This approach is based on the construction of an intrinsic data manifold as a first step.
The dimension of this data manifold is thereby representative of the state space dimension of the data generating process.

We argue that, after successful training, the LSTM thereby approximates this process. This can be observed in the dynamics of the internal states $\vect{c}_t$ and $\vect{h}_t$,
which, after initialization, converge to the ``physical'' manifold after a few time steps.
This ``physical'' manifold must then be one-to-one with the manifold on which the observed data lives.

The warmup or washout phase can be viewed as a driving of the learned dynamical system with the observed data.
A long enough driving phase leads to synchronization, forcing the internal
states to converge to the right ``phase'' on the physical manifold.
The required time until the two systems synchronize, however, may be arbitrarily long.

We therefore chose to circumvent this process of initializing learned LSTM neural networks by using the
concept of generalized synchronization:
if the learned dynamical system and the data generating process are synchronized, then the cell states $\vect{c}_t$ and hidden states $\vect{h}_t$  are a function
on the data manifold $\manifold$: a function that can be learned in a data-driven way. Here this is done with geometric harmonics. 
Given a short input sequence $\utSeq$, one can therefore find a mapping $\utSeq \mapsto \manifold$ and learn the function $\manifold \mapsto \left\{\cZero(\utSeq), \hZero(\utSeq)\right\}$,
providing initial conditions that are consistent with the input sequence $\utSeq$.
The required length of the input sequence depends on the dimension of the data manifold (and thus of the data generating process):
using the Takens embedding theorem, $2n+1$ time steps are sufficient, with $n$ being the dimension of the manifold.

Having learned the data manifold simplifies the nonlinear system identification task by transforming a problem of partially observed data (only $u_t$ is observed)
into a problem where the number of observed variables corresponds (accounting for the Whitney and Takens embedding theorems) to the state space dimension of the data generating process.

Here, we propose an approach for consistent initialization of an
LSTM model after it has been trained successfully.
An exciting issue worth investigating in the future is the proper initialization of LSTM neural networks \textit{during the training phase}, which still poses an open problem.

Having obtained the physical manifold of the internal states
allows us to further optimize the LSTM
by imposing constraints on the dynamics off manifold!
That is, in the case discussed above we observed that the $\vect{c}_t$ quickly converge onto
a two-dimensional manifold.
We can now impose desired stability properties, that is, strengthen the attractivity of the dynamics transverse
to the physical manifold, by regularizing the Jacobian~\cite{hoffman19_robus_learn_with_jacob_regul,pan18_long_time_predic_model_nonlin},
i.e. the eigenvalues corresponding to transverse directions, using automatic differentiation.

Another issue worth investigating in the future is that discrete-time dynamical systems may have multiple pre-images for a given state~\cite{gicquel98_noninvertability, RICOMARTINEZ20002417, 064e69a6792d4ebdad65f26dbf31694e, NEURIPS2018_4ff6fa96, jaeger2017controlling}. This should also be observable for LSTM neural networks!

A few final remarks: (a) Suppose that occasional measurements of the ``other'', hidden variable $v(t)$ are available from time to time; not necessarily as a time series, but as individual sporadic measurements.
Then, with enough such measurements, since $v(t)$ is also a function over the intrinsic low-dimensional manifold, this function can be imputed through geometric harmonics, and the full system state can thus be available.
(b) Since every system observable is a function over the intrinsic manifold, so are also the time derivatives $du/dt$ and $dv/dt$; so given sufficient (even sporadic) measurements, {\em one could identify also the full dynamical system!}

A separate (and for parametrically dependent predictions, very important) issue, discussed in detail in \cite{rico-martinez92_discr_vs,hudson90_nonlin_signal_proces_system_ident,ANDERSON1996S751, chapter_bulsari} is that while the short-term accuracy of discrete time models of continuum time systems can be more than satisfactory, the long-term dynamics and bifurcations are generically {\em simply wrong}. Indeed, a discrete time model does not have limit cycles, but rather, invariant circles, whose rotation number is a fractal function (a "devil's staircase") of system parameters or of the time step. Period doublings and turning points of invariant circles are not generic in one-parameter diagrams, while they very much are generic for limit cycle solutions. It becomes then important {\em not} to use LSTMs if correct bifurcations of the system dynamics are expected to be captured by the model.

Finally: all we discussed here for LSTM initialization also holds for reservoir computing; we are currently working on demonstrating this. 


\section{Methods}
\label{sec:methods}

\subsection{Training}
\label{sec:training}

A total of 400 trajectories, obtained from different initial conditions was used for training, and additional 50 trajectories were used for validation and additional 50 trajectories for testing.
The initial conditions are drawn uniformly as $u_0\in\left[0, 2\right]$ and $v_0\in\left[0, 3\right]$,
and the resulting trajectories are sampled as explained in the main text.
The neural networks are optimized using teacher forcing~\cite{Williams1989, Bengio2015} and the mean-squared error at each time step as a loss function.
The internal states $\vect{c}_0$ $\vect{h}_0$ are initialized as zero during training.
As optimizer, Adam~\cite{kingma2017adam} with PyTorch's default hyperparameters was used~\cite{NEURIPS2019_9015}.
Each model was trained for 1000 epochs,
a batch size of 128, and initial learning rate of $5\cdot 10^{-3}$.
The learning rate was halved when the training error did not decrease for 25 epochs.



\subsection{Diffusion Maps}
\label{sec:dmaps}
Diffusion maps parametrization can be used for dimensionality reduction of a finite data set,
$\mat{X} = \{\vect{x}_i\}^{N}_{i=1}$,
where the $\vect{x}_i \in \mathbb{R}^{m}$ are sampled from a manifold $M$~\cite{Coifman2006}.
We note, before starting, that what is accomplished here through diffusion maps can also be accomplished through Gaussian process modeling; we will not demonstrate this here. 

The first step of diffusion maps involves the construction of a random walk on the data set.
This is achieved by the means of an affinity matrix $\mat{K} \in \mathbb{R}^{N \times N}$
encoding the \textit{connectivity} between the points in $\mat{X}$.
The entries of this matrix $\mat{K}$ are computed in terms of a kernel, e.g. a Gaussian kernel,
\begin{equation}
    K_{ij}=\exp\left(-\frac{\norm{\vect{x}_i-\vect{x}_j}^2}{2\epsilon}\right) 
    =K\left(\vect{x}_i, \vect{x}_j\right)
    \label{eq:Kernel},
\end{equation}
where $\norm{\cdot}$ is the ``appropriate norm'' for the observations~\cite{Coifman2006}.
Here, we consider only the $L^2$ norm.
The hyperparameter $\epsilon>0$ regulates the rate of decay of the kernel:
for small values of $\epsilon$, only points that are close to each other are considered as connected
in $\mat{K}$, since distant points will have $K_{ij}\approx0$.

The diffusion maps algorithm is based on the convergence of the normalized graph Laplacian on the
data to the Laplace-Beltrami operator on the manifold $M$,
as the number of points $N \to \infty$ and $\epsilon \to 0$.
But assuming that the data was obtained from non uniformly sampled points a subtle
normalization needs to be done to recover the Laplace-Beltrami operator.
To this end, we define a diagonal matrix $\mat{P} \in \mathbb{R}^{N \times N}$ with entries
\begin{equation}
  P_{ii}=\sum^{N}_{j=1}K_{ij}
  \label{eq:P_computation}
\end{equation}
and compute the normalized affinity matrix
\begin{equation}
  \mat{\widetilde{K}}=\mat{P}^{-\alpha}\mat{K}\mat{P}^{-\alpha}.
  \label{eq:Normalization_density}
\end{equation}
The parameter $\alpha$ controls the effect of the density.
For $\alpha =0$ the influence of the density is maximal and the approximation of the
Laplace-Beltrami operator is valid only in the case
of uniform sampling~\cite{coifman08_graph_laplac_tomog_from_unknow_random_projec}.
In the case of non-uniform sampling $\alpha=1$ factors out the density effect and
the Laplace-Beltrami operator is obtained.
Another normalization is applied,
\begin{equation}
    {D}(\vect{x}_i,\vect{x}_j) = \frac{{\widetilde{K}}(\vect{x}_i,\vect{x}_j)}{\sum_{j=1}^{N} {\widetilde{K}}(\vect{x}_i,\vect{x}_j)},
\end{equation}
leading to the construction of $\mat{D}$, a row-stochastic or Markovian matrix.
It can be shown that the eigendecomposition of $\mat{D}$
has a complete set of real eigenvectors $ \vect{\phi}^{(i)}$ and
eigenvalues $\lambda_{i}$~\cite{berry13_time_scale_separ_from_diffus},
\begin{equation}
    \label{eq:Eigendecomposition}
    \mat{D}\vect{\phi}^{(i)} = \lambda{_i}\vect{\phi}^{(i)}.
\end{equation}
A non-linear parametrization of the original data set $\mat{X}$ is given in terms of those
computed eigenvectors.
Proper selection of the independent/non-harmonic leading eigenvectors gives a set
of latent variables $\embeddingMatrix=\left\{\vect{\phi}^{(1)}, \dots, \vect{\phi}^{(d)}\right\}$ that span the \textit{intrinsic} geometry of the manifold $M$ from which the original
data set was sampled~\cite{dsilva18_parsim_repres_nonlin_dynam_system}.
If the number of those independent eigenvectors $d$ is smaller than the number of the
original variable dimensions $m$ then the algorithm achieves
dimensionality reduction by revealing a more parsimonious representation of the original data set.\\

Given a data set $\mat{X}$ of short time series windows $u_t$, we use diffusion maps here to obtain, in a data-driven way, a set of reduced latent variables. As hyperparameters, we use $\alpha=0$ and $\epsilon$ as the median of all pairwise distances, given that the choice of $\alpha$ did not qualitatively alter the diffusion map results.

\subsection{Nystr\"om Extension}
\label{sec:Nystrom Extension}
The Nystr\"om extension finds numerical approximations to
eigenfunction problems~\cite{nystroem28_ueber_prakt_aufloes_von_lin, fowlkes01_effic_nystr} of the form
\begin{equation}
    \label{eq:NystromIntegral}
    \int_{a}^b {W}(\vect{x}_i,\vect{x}_j)\vect{\phi}({\vect{x}_j) = \lambda\vect{\phi}(\vect{x_i}}).
\end{equation}
In the context of our paper, Nystr\"om extension is being used as an interpolation scheme for
new unseen data points.
More precisely, given a sample point $\vect{x}_{new} \notin \mat{X}$, Nystr\"om extension computes $\vect{\Phi}_{new}$ for this point with the algorithm described below.
The first step is to compute the distance,
in our case Euclidean distance, between this new point,
$\vect{x}_{new}$ and all the preexisting points in the data set $\mat{X}$,
\begin{equation}
\label{eq:Kernel_nystrom}
{K}(\vect{x}_{new},\vect{x}_j) = \exp \left (-\frac{|| \vect{x}_{new} - \vect{x}_j||^2}{2\epsilon} \right).
\end{equation}
The same density normalization parameter $\alpha$ is being used as before,
\begin{equation}
     {\widetilde{K}}(\vect{x}_{new},\vect{x}_j) = \frac{{K}(\vect{x}_{new},\vect{x}_j)}{{p}(\vect{x}_{new})^{\alpha}\vect{p}(\vect{x}_j)^{\alpha}}
\end{equation}
with $p(\vect{x}_{new}) = \sum_{j=1}^N {{K}}(\vect{x}_{new},\vect{x}_j)$ being just a scalar value and  $\vect{p}(\vect{x}_{j}) = \sum_{i=1}^N {K}(\vect{x}_j,\vect{x}_i)$ being a N-dimensional vector.
The kernel $W$ is then defined as
\begin{equation}
    {W}(\vect{x}_{new},\vect{x}_j) = \frac{{\widetilde{K}}(\vect{x}_{new},\vect{x}_j)}{\sum_{j = 1}^N {\widetilde{K}}(\vect{x}_{new},\vect{x}_j)}.
\end{equation}
Using this expression, the value of the $\beta$-th reduced coordinate is given by
\begin{equation}
\label{eq:Nystrom_Expression}
    {\phi}^{(\beta)} (\vect{x}_{new}) = \frac{1}{\lambda_{\beta}} \sum_{j=1}^N {W}(\vect{x}_{new},\vect{x}_j)\phi^{(\beta)}(x_{j})
\end{equation}
where $\phi^{(\beta)}(x_{j})$ is the $j$-th component of the $\beta$-th eigenvector $\vect{\phi}^{(\beta)}$ and $\lambda_{\beta}$ is the $\beta$-th eigenvalue~\cite{schoelkopf98_nonlin_compon_analy_as_kernel_eigen_probl, williams01_using_nystr}.

In our work, Nystr\"{o}m extension is used to map new ambient space points, windows of $u_t$,
to the reduced diffusion maps coordinates (also called restriction).

\subsection{Double Diffusion Maps - Geometric Harmonics}
\label{sec:GH}
Given a (possibly vector-valued) function $\mat{F}$ sampled on some points $\xdata$ on a manifold $M$,
geometric harmonics aims to extend the function in a neighborhood for
$\vect x_{new} \notin \xdata$~\cite{Coifman2006}.
Here, we use a \textit{slightly twisted} version of geometric harmonics
to perform interpolation of the function $\mat{F}$ on the reduced coordinates $\embeddingMatrix$ discovered
by diffusion maps (cf. \sect~\ref{sec:dmaps}).
In our case, given the non-harmonic eigenvectors computed during the dimensionality reduction step,
we aim to \textit{write} $\mat{F}$ in terms of those reduced coordinates
$
    \Phi_{i,j}
    =
    \phi_{i=1,\ldots,N}^{(j=1,\ldots,d)}
$.
Given columns of $\embeddingMatrix$, the $d$ non-harmonic eigenvectors, we cannot map directly to the function $\mat{F}$,
since we discarded the harmonic eigenvectors.
However, computing a second round of diffusion maps on the reduced diffusion maps coordinates $\embeddingMatrix$
allows us to construct a basis of functions with which we can map from the reduced coordinates $\embeddingMatrix$ to any function $\mat{F}$ defined on the ambient space coordinates.


As in the $1^{st}$ round of diffusion maps, the first step here is to compute an affinity matrix
\begin{equation}
\label{eq:output_kernel}
        \secondKernel_{i,j}
        =
        \secondKernel(\vect{\phi}_i,\vect{\phi}_j)=\exp \left (-\frac{\norm{\vect{\phi}_i - \vect{\phi}_j}^2}{2\epsilon^{\star}} \right).
\end{equation}
Since it is symmetric and positive semidefinite,
$\mat\secondKernel$ has a set of orthonormal vectors $\psi^{(1)},\psi^{(2)},\ldots\psi^{(N)}$ and non-negative eigenvalues ($\sigma_1 \geq \sigma_2 \geq\cdots\geq \sigma_N \geq 0$).
Those eigenvectors are used as a basis in which we can project and subsequently extend any
function $\mat{F}$.
For some $\delta >0$ we consider the set of truncated eigenvalues
$S_{\delta} = \{\alpha: \sigma_{\alpha}> \delta\sigma_{1}\}$
(where we can select $\delta$ s.t.
$d<\norm{S_{\delta}}<N$).
In the corresponding truncated set of eigenvectors we project the values of $F_{i}^{(\ell)}$, packed in the matrix $\mat F$, as
\begin{equation}
    \mat F
  \approx
    P_{\delta}\mat F 
  \equiv
    \mat\Fproj
  =
      \sum_{\alpha\in{S_{\delta}}} 
      \psi^{(\alpha)} \cdot
      \left(
        \mat F^T \cdot \psi^{(\alpha)}
      \right) ^ T.
  \label{eq:Geometric_Harmonics_Projection}
\end{equation}
The extension of $\mat \Fproj$ to $\phinew \notin \embeddingMatrix$ is defined by
\begin{equation}
    \label{eq:Geometric_Harmonics_Extension}
    \vect{\Fproj}_\new (\phinew) = 
    \sum_{{\alpha\in{S_{\delta}}}} 
    \psi^{(\alpha)}_\new \cdot
    \left(
    \mat \Fproj^T \cdot \psi^{(\alpha)}
  \right) ^ T.
\end{equation}
with
\begin{equation}
    \label{eq:Geometric_Harmonics_functions}
    \psi^{(\alpha)}_\new
    =
    \sigma^{-1}_{\alpha}
    \sum_{i=1}^{N}
    \secondKernel(\phinew,\vect{\phi}_i)
    \cdot
    \psi^{(\alpha)}\left(\vect{\phi}_i\right)
\end{equation}
and where the scalar $\psi^{(\alpha)}\left(\vect{\phi}_i\right)$ is the $i$-th component of the diffusion maps eigenvector $\vect{\psi}^{(\alpha)}$.
%
%
It is worth noting that using a truncated set $S_{\delta}$ is important to circumvent the numerical instabilities arising in Eq.~\eqref{eq:Geometric_Harmonics_functions} when $\sigma_{\alpha} \to 0$.

Using geometric harmonics this way,
we can estimate the values of $\mat{F} = \left[
    \vect{c}_t, \vect{h}_t\right
]$
for unseen points $\left(\phi^{(1)}_{{new}}, \phi^{(2)}_{{new}}\right)$.
Here, $\phi^{(\beta)}_{{new}}$, $\beta \in \left\{1, 2\right\}$, is obtained by Nystr\"{o}m extension on time series windows
of $u_t$ (here, of length 5).

\begin{figure}
    \centering
    \includegraphics[width=\textwidth]{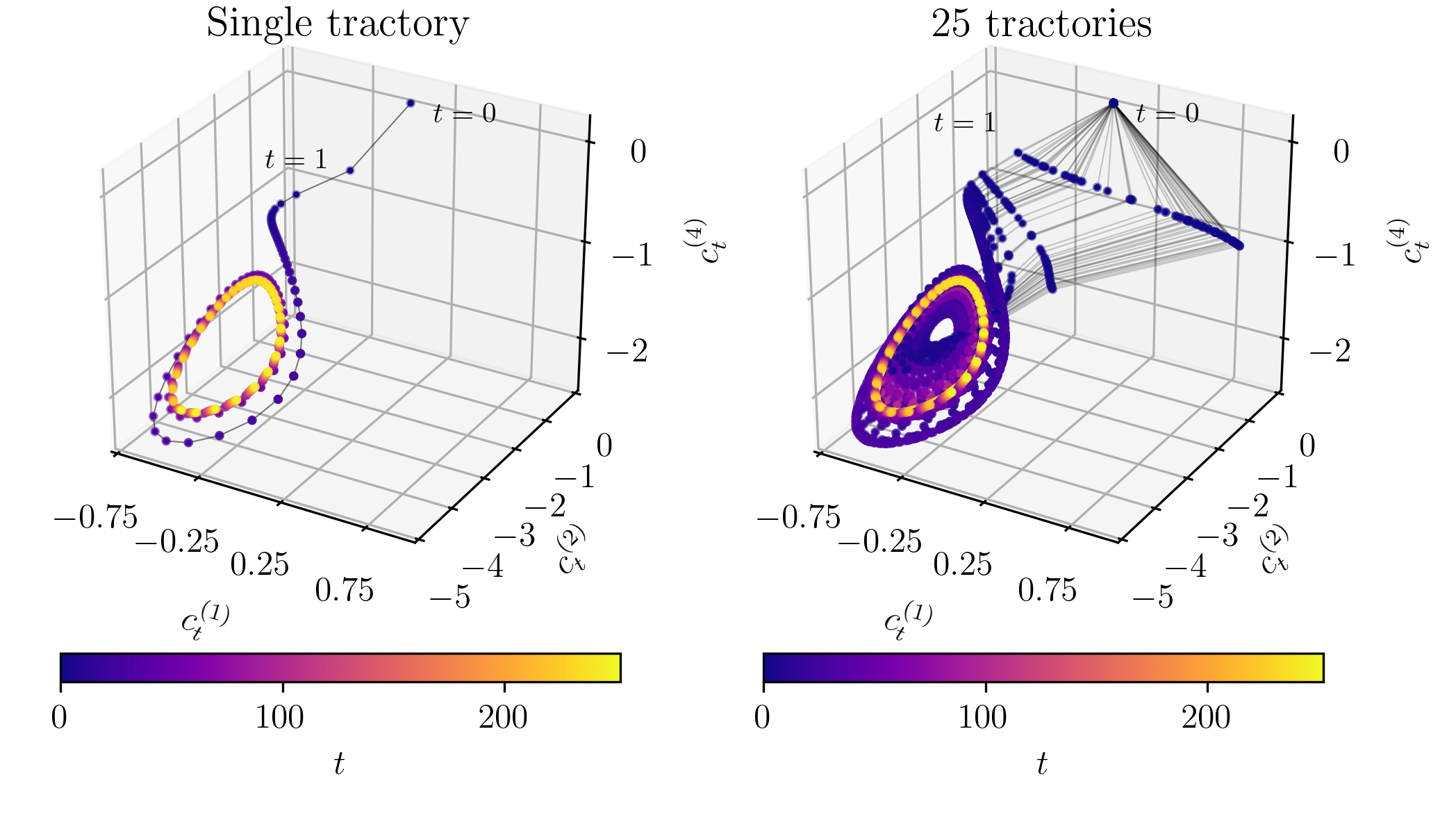}
    \caption{The cell states $\vect{c}_t$ obtained by integration without warmup, projected onto the three variables $c_t^{(1)}$, $c_t^{(2)}$ and $c_t^{(4)}$. On the left, the projection of the $\vect{c}_t$ values of only a single trajectory is shown, whereas on the right the values of 25 trajectories are depicted. The color corresponds to the time step $t$.}
    \label{fig:dmaps_ct_wo_warmup_124}
\end{figure}

\begin{figure}
    \centering
    \includegraphics[width=\textwidth]{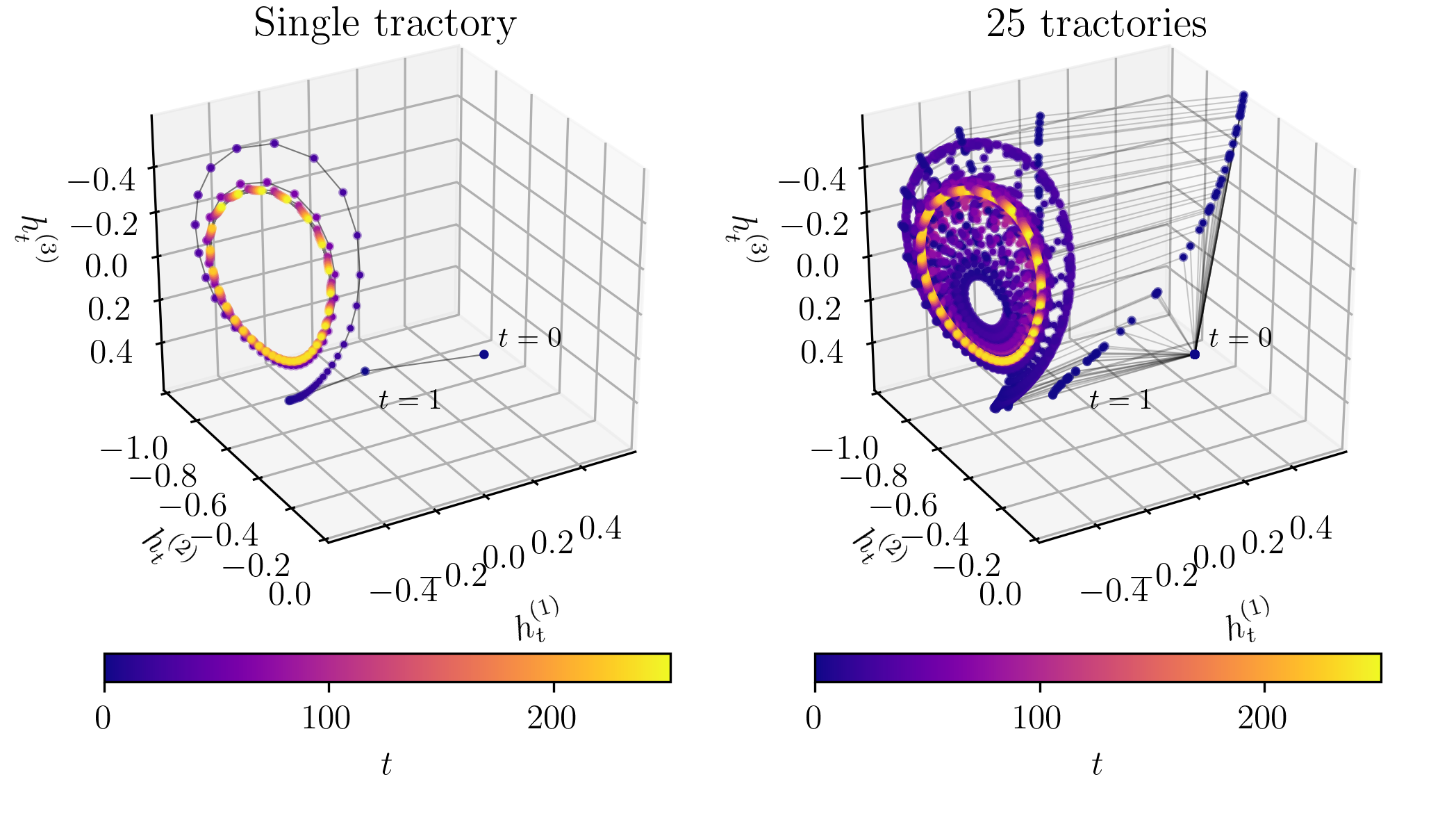}
    \includegraphics[width=\textwidth]{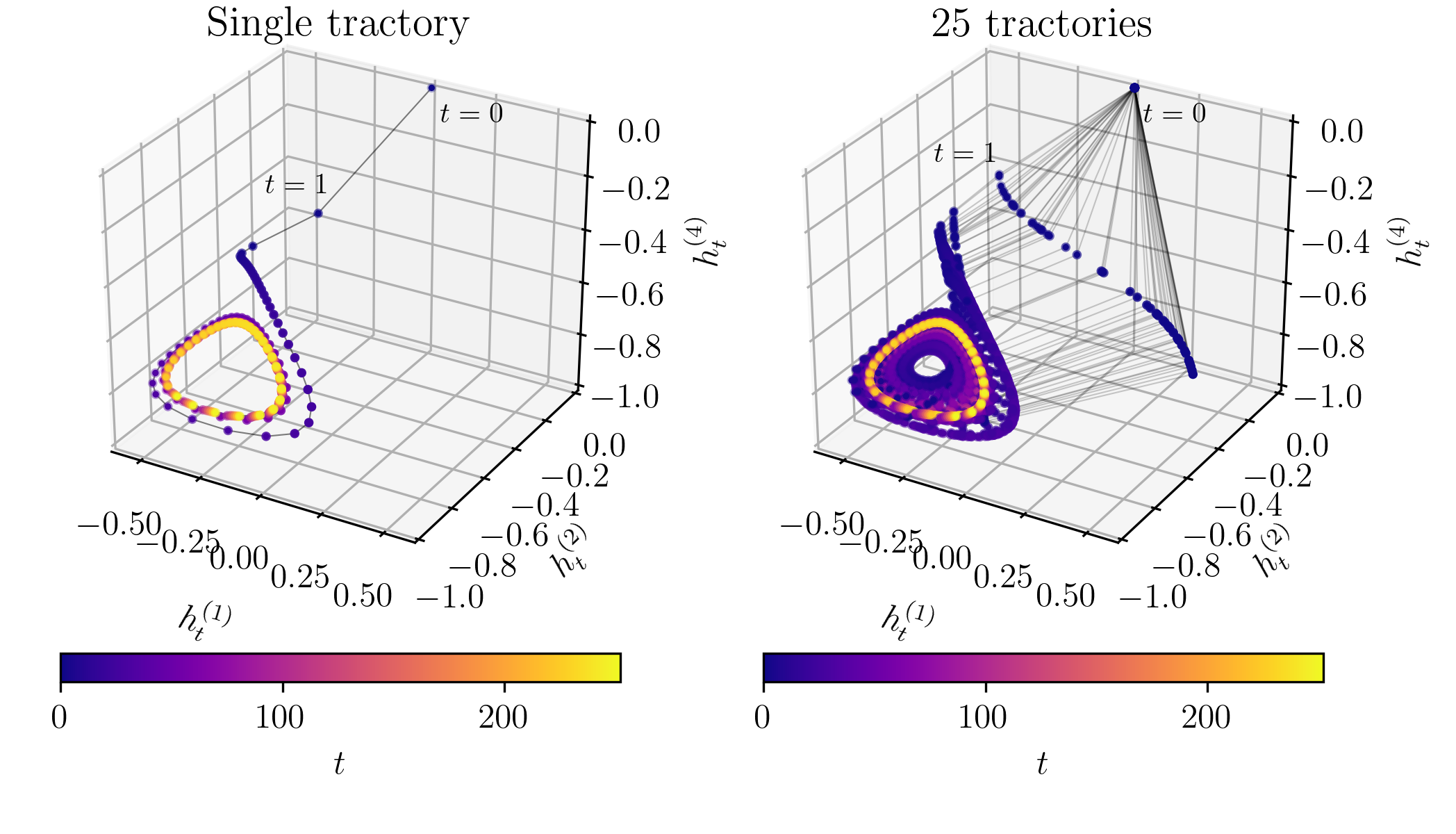}
    \caption{The hidden states $\vect{h}_t$ obtained by integration without warmup, projected onto the three variables $h_t^{(1)}$, $h_t^{(2)}$ and $h_t^{(3)}$ (top) and $h_t^{(1)}$, $h_t^{(2)}$ and $h_t^{(4)}$ (bottom). On the left, the projection of the $\vect{h}_t$ values of only a single trajectory is shown, whereas on the right the values of 25 trajectories are depicted. The color corresponds to the time step $t$.}
    \label{fig:dmaps_ht_wo_warmup}
\end{figure}

{\bf Acknowledgements}: This work was partially supported by the US Department of Energy, the Army Research Office through a MURI, and the DARPA ATLAS program.

\bibliography{lit.bib}
\bibliographystyle{unsrt}

\end{document}